\documentclass[conference]{IEEEtran}
\IEEEoverridecommandlockouts
\usepackage{cite}
\usepackage{amsmath,amssymb,amsfonts}
\usepackage{algorithmic}
\usepackage{graphicx}
\usepackage{textcomp}
\usepackage{xcolor}
\def\BibTeX{{\rm B\kern-.05em{\sc i\kern-.025em b}\kern-.08em
    T\kern-.1667em\lower.7ex\hbox{E}\kern-.125emX}}

\usepackage{multirow}
\usepackage{url}

\begin{document}

\title{Survey of Visual-Semantic Embedding Methods \\for Zero-Shot Image Retrieval
}

\author{\IEEEauthorblockN{Kazuya Ueki}
\IEEEauthorblockA{\textit{Department of Information Science} \\
\textit{Meisei University}\\
Tokyo, Jappan \\
kazuya.ueki@meisei-u.ac.jp}
}

\maketitle

\begin{abstract}
Visual-semantic embedding is an interesting research topic because it is useful for various tasks, such as visual question answering (VQA), image-text retrieval, image captioning, and scene graph generation.
In this paper, we focus on zero-shot image retrieval using sentences as queries and present a survey of the technological trends in this area.
First, we provide a comprehensive overview of the history of the technology, starting with a discussion of the early studies of image-to-text matching and how the technology has evolved over time.
In addition, a description of the datasets commonly used in experiments and a comparison of the evaluation results of each method are presented.
We also introduce the implementation available on github for use in confirming the accuracy of experiments and for further improvement.
We hope that this survey paper will encourage researchers to further develop their research on bridging images and languages.
\end{abstract}


\section{Typical Approaches for Visual-Semantic Embedding}
\label{sec:approaches}

\begin{figure*}[h]
 \centering
 \includegraphics[width=130mm]{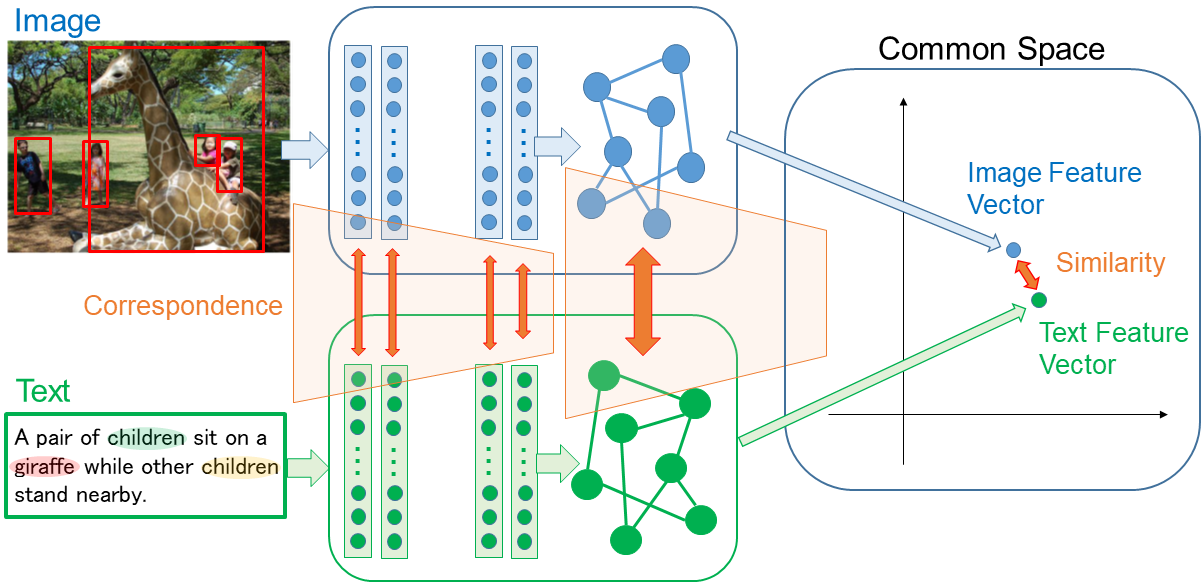}\\
 \vspace{-2mm}
 \caption{General architecture for visual-semantic embedding}
 \label{fig:architecture}
\end{figure*}

\begin{figure*}[h]
    \centering
	\includegraphics[width=125mm]{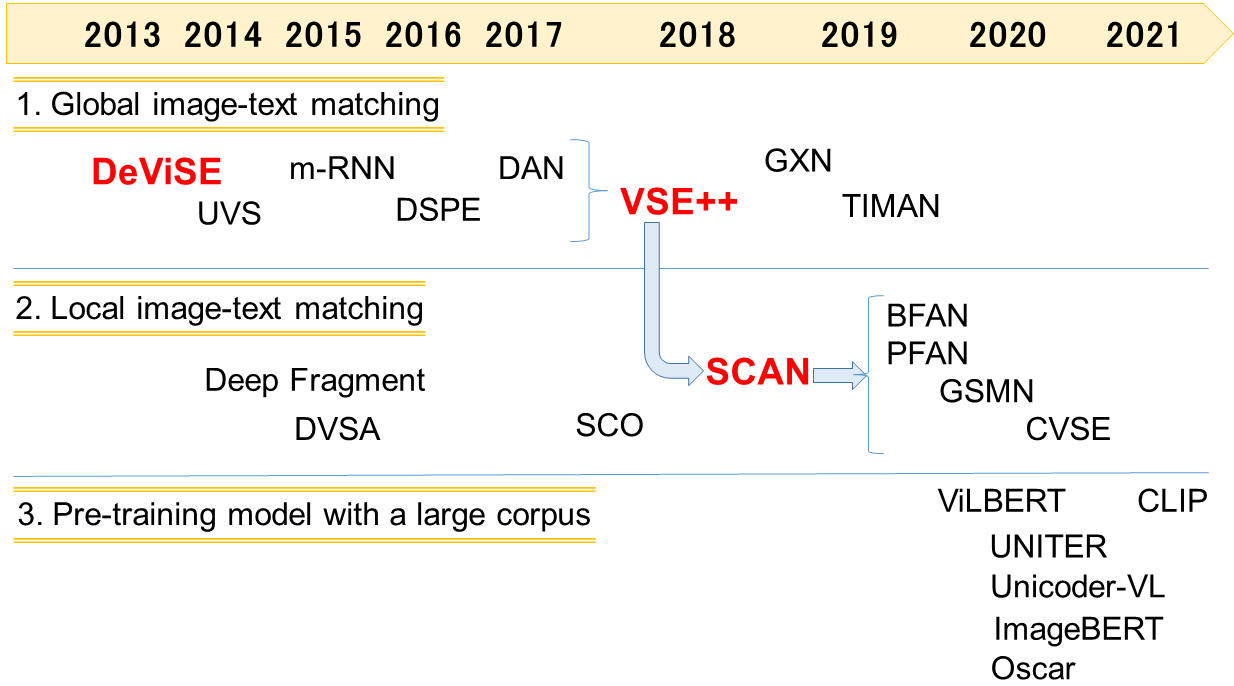}\\
	\vspace{-2mm}
	\caption{History of visual-semantic embedding methods}
	\label{fig:history}
\end{figure*}

A typical solution for visual-semantic embedding is to map images and language to a common embedding space, as shown in Fig.~\ref{fig:architecture}.
In recent years, various methods have been proposed; these methods can be roughly classified into the following three categories:
\begin{enumerate}
\item 
A method for computing the similarity between images and language by training a deep neural network to extract global representations of images and language.
\item 
A method for training local correspondences between salient regions in an image and words in a sentence using a deep neural network and calculating their similarity.
\item 
A method that utilizes pre-trained models employing a large corpus of images and text
\end{enumerate}
The history of each method is shown in Fig.~\ref{fig:history}.
We introduce some of the representative methods below.

\subsection{Methods for Global Image-Text Matching}
\label{subsec:global}

In the method of mapping global representations, features from the whole image and features from the text are transformed into a common space, and their similarity is measured.

An early method for mapping images and language into a common space is deep visual-semantic embedding (DeViSE) \cite{DeViSE}, proposed by Frome et al.
DeViSE embeds images and their corresponding words in the same space and trains them so that their similarity is high.
Image features were extracted from convolutional neural networks (CNNs) trained with ImageNet, and word features were extracted from models trained with the skip-gram language model.
For $z_i = M x $, which is the feature $x$ extracted by inputting image $i$ into the CNN and multiplying by the transformation matrix $M$, for $z_t$, which is the feature obtained from word $t$, the cosine similarity is calculated as follows:
\begin{equation}
 {\rm sim}(i, t) = \frac{z_i \cdot z_t}{\|z_i\|_2 \|z_t\|_2}.
\end{equation}
For every negative example caption $\hat{t}$ for image $i$, the matrix $M$ that linearly transforms the image features into the word feature space is trained using the following hinge rank loss:
\begin{equation}
L(i, t) = \sum_{\hat{t}} \max [0, \alpha - {\rm sim}(i, t) + {\rm sim}(i, \hat{t})] \notag, 
\end{equation}
where $\alpha$ is the margin.

Unifying visual-semantic embeddings (UVS) \cite{UVS}, proposed by Kiros et al., uses a CNN trained on ImageNet (AlexNet \cite{AlexNet}) for image feature extraction and LSTM for text feature extraction.
Let $x,y$ be the features of the image and text, respectively, and $M_x$ and $M_y$ be the transformation matrices, respectively. Then, the vectors in the embedded common space can be denoted as $z_i=M_x x$ and $z_t=M_y y$.
The similarity between image data $i$ and text data $t$, ${\rm sim}(i,t)$, is calculated using the cosine similarity of vectors $z_i, z_t$ in the common space.
The loss function is a bi-directional hinge rank loss, as shown below:
\begin{align}
L(i,t) =& \sum_{\hat{t}} \max \{ 0,\alpha - {\rm sim}(i,t) + {\rm sim}(i,\hat{t})\} \notag \\
& + \sum_{\hat{i}} \max \{ 0,\alpha - {\rm sim}(i,t) + {\rm sim}(\hat{i},t)\} \notag,
\end{align}
where $\hat{i}$ and $\hat{t}$ are data unrelated to $t$ and $i$, respectively.

A multimodal recurrent neural network (m-RNN) \cite{m-RNN} has also been proposed; it introduces a multimodal layer to a recurrent neural network (RNN) and has a mechanism to input image features obtained from the CNN.

Deep structure-preserving embedding (DSPE) \cite{DSPE} has also been proposed; it includes a constraint that preserves the correspondence between different modalities (image and text) even when training the correspondence between the same modality.

Improved visual semantic embedding (VSE++) \cite{VSE++} is an improved version of the UVS method described above, with the following changes to the loss function:
\begin{align}
L(i,t) =&  \max_{\hat{t}} \biggl[
 \max \{ 0,\alpha - {\rm sim}(i,t) + {\rm sim}(i,\hat{t})\} \biggr]
 \notag \\
& + \max_{\hat{i}} \biggl[ \max \{ 0,\alpha - {\rm sim}(i,t) + {\rm sim}(\hat{i},t)\} \biggr] \notag .
\end{align}
UVS sums over all the data, while VSE++ includes only the data with the largest loss at each data point.
This allows us to utilize hard negatives and has been shown to significantly improve accuracy, regardless of the image feature extraction method used, such as VGG or ResNet.

A generative cross-modal feature learning framework (GXN) \cite{GXN} has also been proposed; it incorporates generative processes into the training of cross-modal feature embedding.

Dual attention networks (DANs) \cite{DAN}, which aim to capture the detailed interactions between images and text, incorporate attention mechanisms in both images and text, and experiments have shown their effectiveness.
DANs involve multiple steps to collect essential information from both modalities, focusing on specific regions of the image and words in the text.

Text-image modality adversarial matching (TIMAM) \cite{TIMAM}, which uses adversarial correspondence, has also been proposed to train modality-invariant feature representations.
The paper also describes the successful inclusion of BERT as a language model.

\subsection{Methods for Local Image-Text Matching}
\label{subsec:local}

Methods for extracting global representations of images and text are unable to find the relationship between objects in an image and words in a sentence, which limits the accuracy of image-text matching.
For this reason, many methods have been proposed to measure relevance by mapping parts of an image or a sentence and aggregating their similarities.

Karpathy et al. were the first to locally map images to text by optimizing the correspondence between the most similar image regions and word pairs with deep fragment embeddings \cite{Deep_Fragment}.

Deep visual-semantic alignments (DVSAs) \cite{DVSA} estimate the latent alignment between regions in an image obtained by region CNN (R-CNN) and words in a sentence obtained by a bidirectional recursive neural network (BRNN).
This allows us to deal with sentences whose meaning is interpretable and whose length is not fixed.

A method called semantic concepts and order (SCO) \cite{SCO} has also been proposed.
This method improves the image representation by training semantic concepts and then organizing them in the correct semantic order.

In the above methods, the similarity was calculated for all possible pairs without focusing on regions in the image or words in the sentence.
In contrast, the stacked cross attention network (SCAN) \cite{SCAN}, a method to improve interpretability by focusing on regions of an image or words in a sentence using the bottom-up attention model \cite{Bottom-up}, has been proposed.
SCAN has been used as a baseline for many methods and has led to technological developments since its proposal.
Examples include the bidirectional focal attention network (BFAN) \cite{BFAN} and the position focused attention network (PFAN) \cite{PFAN}.
In BFAN, irrelevant image regions and words cause deterioration in the correspondence between images and text; thus, they are removed.
PFAN trains the correspondence between images and text using the positions of objects in the images as cues.

Training coarse correspondences based on object co-occurrence statistics does not allow for fine phrase correspondences.
To train more detailed correspondences, a graph structured matching network (GSMN) \cite{GSMN}, which models objects, relationships, and attributes as structured phrases through node- and structure-level correspondences, has been proposed and demonstrated greatly improved accuracy.

Scene concept graph (SCG) \cite{SCG}, which extends visual concepts and enhances image representation using image scene graphs as external knowledge, is another effective method.
Another method, consensus-aware visual-semantic embedding (CVSE) \cite{CVSE}, is inspired by the SCG method and is the first attempt to map images to text using common sense information.
CVSE utilizes consensus information by computing the statistical co-occurrence correlations between the semantic concepts from an image caption corpus and expanding the constructed concept correlation graph to generate consensus-aware concept (CAC) representations.

\subsection{Methods Using Pre-trained Models Employing a Large Corpus of Images and Languages}
\label{subsec:pretrained}

\begin{figure}[t]
    \fboxsep=3pt
    \centering
    \small
	\fbox{\includegraphics[width=80mm]{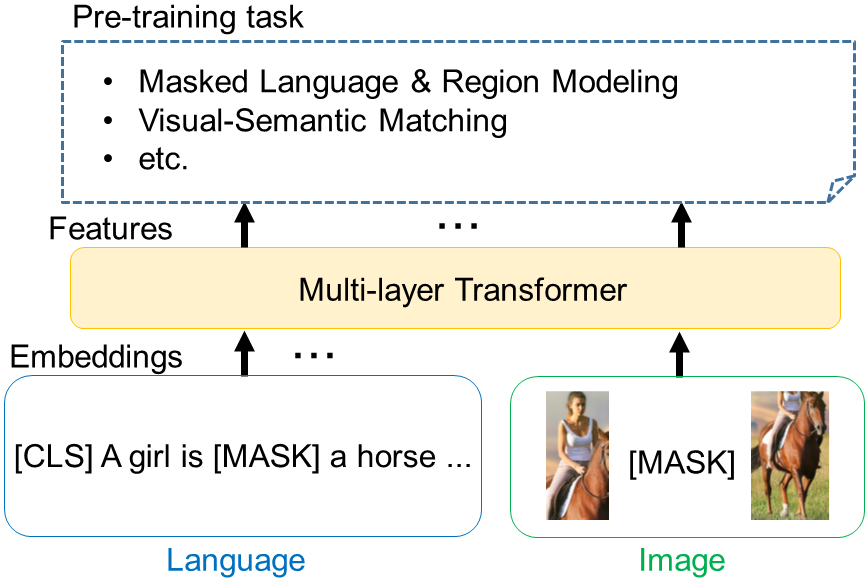}} \\
	Methods for treating images and text as a single sequence \\
	\vspace{3mm}
	\fbox{\includegraphics[width=80mm]{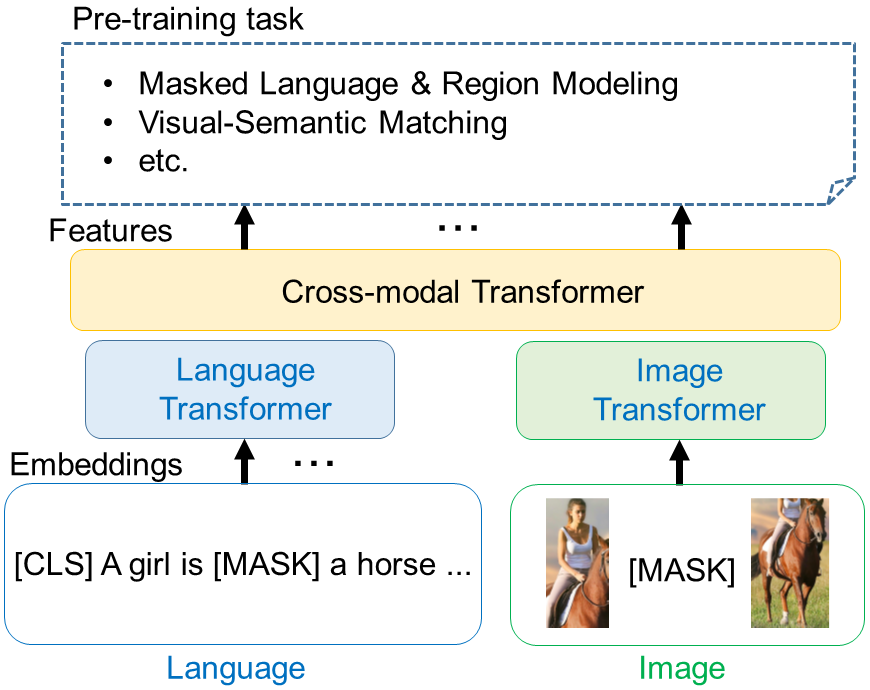}} \\
	Methods for treating images and text as separate sequences \\
	\vspace{1mm}
	\caption{General architecture of methods that utilize pre-trained models employing a large corpus of images and text}
	\label{fig:Transformer}
\end{figure}

Currently, because of the availability of large datasets of image and language pairs and the success of large pre-trained models such as transformers \cite{transformer} and BERT \cite{BERT} in the field of natural language processing, image retrieval using pre-trained models is becoming mainstream, as shown in Fig.~\ref{fig:Transformer}.
Pre-training models are mostly based on multi-layer transformers, and typical datasets used include Flickr30k \cite{Flickr30k}, MSCOCO \cite{MSCOCO}\cite{COCO_Captions}, Conceptual Captions \cite{Conceptual_Captions}, SBU Captions \cite{SBU_captions}, and graph question answering \cite{GQA}.
In the following, we introduce several methods that have been proposed in recent years.

Vision and language BERT (ViLBERT) \cite{ViLBERT}, which extends the BERT model to a multimodal model of images and text, can train common task-independent representations of images and natural language.
In ViLBERT, the image and text inputs are processed in separate sequences by the transformer, after which the cross-modal relationship is trained by the co-attentional transformer, as shown in the lower part of Fig.~\ref{fig:Transformer}.
Experiments show that it is possible to achieve higher accuracy than when  images and text as a single sequence and training them with a single BERT.
The authors used the Conceptual Captions dataset to pre-train ViLBERT and then fine-tuned it for tasks such as VQA, visual sense reasoning (VCR), and image retrieval to achieve the highest accuracy.

Unicoder-VL \cite{Unicoder-VL} is a universal encoder that aims to pre-train common visual and semantic representations using a large image caption dataset.
Language and images are input to a single transformer model with multiple sequential encoders, which train a common embedding.
The following three tasks were used for pre-training:
\begin{itemize}
\item Masked Language Modeling (MLM): Predicting masked input token
\item Masked Object Classification (MOC): Predicting features of masked objects
\item Visual-Linguistic Matching (VLM): Predicting whether image and text pairs are relevant
\end{itemize}

ImageBERT \cite{ImageBERT}, a transformer-based model trained on multiple tasks simultaneously, achieves high retrieval accuracy by crawling a large number of image-text pairs from the Web for pre-training and fine-tuning them in the image retrieval and text retrieval tasks.
The four tasks used in the training are as follows:
\begin{itemize}
  \setlength{\parskip}{0cm} 
  \setlength{\itemsep}{0cm} 
\item Masked Language Modeling (MLM): Predicting masked input token
\item Masked Object Classification (MOC): Predicting labels of masked objects
\item Masked Region Feature Regression (MRFR): Predicting features for object locations
\item Image Text Matching (ITM): Predicting whether image and text pairs are relevant
\end{itemize}

UNiversal Image-TExt Representation (UNITER), which pre-trains BERT models for four tasks using a large database of image/text pairs, has also been proposed.
The following four tasks are used for pre-training:
\begin{itemize}
 \item Masked Language Modeling (MLM): Predicting masked input token
 \item Image Text Matching (ITM): Predicting whether image and text pairs are relevant
 \item Word-Region Alignment (WRA): Training to optimize the alignment between words and image regions
 \item Masked Region Modeling (MRM): Training to reconstruct the masked image
\end{itemize}
A pre-training task was designed to map visual and language input at both the global (image-text) and local (word-region) levels.

In the above methods, the features of the image domain and the text are concatenated and fed into the model, and the model trains the semantic association between the image and the text in a brute force fashion using self-attention.
In contrast, object-semantics aligned pre-training (Oscar) \cite{Oscar} uses object tags detected in the image as anchor points, which greatly improves the generalizability of the pre-trained models.
The pre-training task performed well with only two loss functions: masked token loss and contrastive loss.
Masked token loss considers object tags as language data and trains the system to mask and recover the input text and object tags.
Contrastive loss takes an object tag as image data, replaces it with a different tag sequence with 50\% probability, and trains it to predict whether the object tag is appropriate.

The contrastive-language-image pre-training (CLIP) \cite{CLIP} proposed by OpenAI has also become a hot topic in the field of image and text retrieval since the beginning of 2021.
CLIP achieves zero-shot, highly accurate image retrieval without fine tuning by pre-training on a large number of text/image pairs (over 400 million).

We have introduced some recent transformer-based methods for image retrieval, but many more methods have been proposed.
In addition to image retrieval, research on transformers in computer vision was presented in \cite{Transformer_Survey}.
Among the methods using transformers, there are methods that treat images and language as one sequence, as typified by UNITER and Oscar, and methods that treat images and language as separate sequences, as typified by VilBERT, but no conclusion has yet been reached as to which is more appropriate.
There is also room to consider whether the region obtained by object detection should be used as the image sequence input to the transformer, or whether the entire image region should be used by a grid-based method.

\section{Dataset}

\begin{table*}[t]
  \centering
  \caption{Comparison of Flickr30k and MSCOCO image retrieval results}
  \small
  \begin{tabular}{l|rrr|r|rrr|r} 
    \hline
    \multirow{2}{*}{Method} & \multicolumn{4}{c|}{Flickr30k} & \multicolumn{4}{c}{MSCOCO} \\
    \cline{2-9}
    & $R@1$ & $R@5$ & $R@10$ & $rSum$ & $R@1$ & $R@5$ & $R@10$ & $rSum$ \\
    \hline
    DeViSE \cite{DeViSE}
                       &  6.7 & 21.9 & 32.7 & 113.1 &  --  &  --  &  --  &  -- \\
    UVS \cite{UVS}     & 16.8 & 42.0 & 56.5 & 251.9 &  --  &  --  &  --  &  -- \\
    m-RNN \cite{m-RNN} & 22.8 & 50.7 & 63.1 & 309.5 & 29.0 & 42.2 & 77.0 & 345.7 \\
    DSPE \cite{DSPE}   & 29.7 & 60.1 & 72.1 & 351.0 & 39.6 & 75.2 & 86.9 & 420.7 \\
    VSE++ \cite{VSE++} & 39.6 & 70.1 & 79.5 & 409.8 & 52.0 &  --  & 92.0 &  -- \\
    GXN \cite{GXN}     & 41.5 &  --  & 80.1 &  --   & 56.6 &  --  & 94.5 &  -- \\
    DAN \cite{DAN}     & 39.4 & 69.2 & 79.1 & 413.5 &  --  &  --  &  --  &  -- \\
    TIMAM \cite{TIMAM} & 42.6 & 71.6 & 81.9 & 415.6 &  --  &  --  &  --  &  -- \\
    \hline
    Deep Fragment \cite{Deep_Fragment} 
                       & 10.3 & 31.4 & 44.5 & 197.5 &  --  &  --  &  --  &  -- \\
    DVSA \cite{DVSA}   & 15.2 & 37.7 & 50.5 & 235.2 & 27.4 & 60.2 & 74.8 & 351.2 \\
    SCO \cite{SCO}     & 41.1 & 70.5 & 80.1 & 418.5 & 56.7 & 87.5 & 94.8 & 499.3 \\
    SCAN \cite{SCAN}   & 48.6 & 77.7 & 85.2 & 465.0 & 58.8 & 88.4 & 94.8 & 507.9 \\
    SCG \cite{SCG}     & 49.3 & 76.4 & 85.6 & 468.7 & 61.4 & 88.9 & 95.1 & 517.6 \\
    BFAN \cite{BFAN}   & 50.8 & 78.4 &  --  &  --   & 59.4 & 88.4 &  --  &  --   \\
    PFAN \cite{PFAN}   & 50.4 & 78.7 & 86.1 & 472.0 & 61.6 & 89.6 & 95.2 & 518.2 \\
    GSMN \cite{GSMN}   & 57.4 & 82.3 & 89.0 & 496.8 & 63.3 & 90.1 & 95.7 & 522.5 \\
    CVSE \cite{CVSE}   & 56.1 & 83.2 & 90.0 & 487.7 & 66.3 & 91.8 & 96.3 & 525.5 \\
    \hline
    ViLBERT \cite{ViLBERT}
                       & 58.2 & 84.9 & 91.5 &  --   &  --  &  --  &  --  &  -- \\
    Unicoder-VL \cite{Unicoder-VL} 
                       & 71.5 & 90.9 & 94.9 & 539.7 & 69.7 & 93.5 & 97.2 & 541.3 \\
    ImageBERT \cite{ImageBERT}
                       & 73.1 & 92.6 & 96.0 & 545.5 & 73.6 & 94.3 & 97.2 & 549.0 \\
    UNITER \cite{UNITER}
                       & 75.6 & 94.1 & 96.8 & 550.9 &  --  &  --  &  --  &  -- \\
    Oscar \cite{Oscar} &  --  &  --  &  --  &  --   & 78.2 & 95.8 & 98.3 & 560.6 \\
    \hline
  \end{tabular}
  \label{tab:recall}
\end{table*}

In this section, we introduce some of the databases commonly used in research related to visual-semantic embedding.

\subsection{Flickr8k/30k}

Databases that are frequently used in training models to generate descriptions (captions) from images include Flickr8k \cite{Flickr8k} and Flickr30k \cite{Flickr30k}.

Flickr8k is a crowdsourced dataset of 8,092 images with five corresponding human-generated captions per image, focusing on people or animals (e.g., dogs) performing some action.
Of the total dataset, 6,000 images are often used as training data, 1,000 images as validation data, and 1,000 images as test data.

Flickr30k is a dataset that extends Flickr8k by annotating each of the 31,783 images with five sentences, in the same way as Flickr8k.
Of the total dataset, 28,000 to 29,783 images are often used as training data, 1,000 as validation data, and 1,000 as test data.

\subsection{MSCOCO}

Microsoft common objects in context (MSCOCO) is a dataset created for object recognition, object detection, and caption generation in images, and it is also frequently used for image-to-language matching and retrieval.
In total, the dataset consists of more than 300,000 images, but the most commonly used are the 123,287 images in the MSCOCO 2014 dataset and the caption data \cite{COCO_Captions}, which contains five descriptions for each image.
The original database consisted of 82,783 training data and 40,504 validation data, but it has become standard practice to use 5,000 of the validation data for testing and the remaining $82,783+40,504-5,000=118,287$ for training data.
Test data for the task of image retrieval are often evaluated by averaging the accuracy of five captions over 1,000 or 5,000 images.

\subsection{Visual Genome}

A well-known dataset that assigns descriptions to regions in an image is Visual Genome \cite{Visual_Genome}, which was created in Li Fei-Fei's lab at Stanford University.
The number of images is 108,077, but the number of labels is overwhelmingly large, with approximately 5.4 million captions assigned to image regions.
In addition, there are approximately 2.8 million labels for attributes of objects in the image, and approximately 2.3 million labels for relationships among objects.

\subsection{Conceptual Captions}

Conceptual Captions \cite{Conceptual_Captions} is a dataset of over three million image and caption pairs published by Google.
The images and captions were collected from the Internet and thus contain a wide range of expressions.
The captions were obtained from the HTML alt attribute associated with each image.
Because of the very large dataset, it is used to train the transformer-based pre-training model shown in Subsection \ref{subsec:pretrained}.

\subsection{SBU Captions}

SBU Captions is a database of one million images with one caption for each image.
The captions are written by humans and filtered to leave only those with at least two nouns, noun--verb pairs, or verb--adjective pairs.
This dataset has been used for research purposes with several tasks, including Google's Show-and-Tell \cite{Show_and_Tell} and Microsoft's UNITER.

\section{Evaluation Criteria}

\begin{table*}[t]
  \centering
  \caption{Publicly available implementation for visual-semantic embedding}
  \small
  \begin{tabular}{l|l} 
    \hline
    Method & URL \\
    \hline
    m-RNN \cite{m-RNN} & \url{https://github.com/mjhucla/mRNN-CR} \\
    VSE++ \cite{VSE++} & \url{https://github.com/fartashf/vsepp} \\
    Bottom-up attention \cite{Bottom-up} & \url{https://github.com/peteanderson80/bottom-up-attention} \\
    SCAN \cite{SCAN} & \url{https://github.com/kuanghuei/SCAN} \\
    PFAN \cite{PFAN} & \url{https://github.com/HaoYang0123/Position-Focused-Attention-Network}, \\
    GSMN \cite{GSMN} & \url{https://github.com/CrossmodalGroup/GSMN} \\
    CVSE \cite{CVSE} & \url{https://github.com/BruceW91/CVSE} \\
    UNITER \cite{UNITER} & \url{https://github.com/ChenRocks/UNITER} \\
    Oscar \cite{Oscar} & \url{https://github.com/microsoft/Oscar} \\
    CLIP \cite{CLIP} &
    \url{https://github.com/openai/CLIP} \\
    \hline
  \end{tabular}
  \label{tab:implementation}
\end{table*}

Typical databases used to evaluate the performance of text-to-image retrieval include Flickr30k \cite{Flickr30k} and MSCOCO \cite{MSCOCO}.

A commonly used evaluation metric for text-to-image retrieval is Recall@K ($K=1,5,10$), where $R@1$, $R@5$, and $R@10$ are the percentage of correct answers contained in the top 1, 5, or 10 ranked images corresponding to the text, respectively.
In addition, the sum of the respective recall values ($rSum$) may be used to indicate the overall mapping performance of text retrieval from images and image retrieval from text, as shown below:
\begin{eqnarray}
 rSum & = & \underbrace{R@1 + R@5 + R@10}_\mathrm{Text~search~with~image~as~a~query} \nonumber \\ \nonumber \\
 & & + \underbrace{R@1 + R@5 + R@10}_\mathrm{Image~search~with~text~as~a~query}. \nonumber
\end{eqnarray}

\section{Evaluation Results}

For the methods shown in Section \ref{sec:approaches}, Table \ref{tab:recall} summarizes the evaluation results for the zero-shot image retrieval presented in the papers for each method.
The databases used for the evaluation are Flickr30k and MSCOCO, which are often compared.
Most of the results in this table were output by the following data division.
\begin{itemize}
 \item {\bf Flickr30k} \\train : validation  : test = 29,783 : 1,000 : 1,000
 \item {\bf MSCOCO} \\train : validation  : test = 123,287 : 113,783 : 1,000
\end{itemize}

The upper part of Table \ref{tab:recall} shows the results of the methods for mapping the global representations of images and languages shown in Subsection \ref{subsec:global}.
From DeViSE, which is the first attempt to convert images and languages into a common space, it can be confirmed that various improvements, such as VSE++, which utilizes hard negatives, and DAN, which uses an attention mechanism, resulted in improved accuracy.
Among them, VSE++ has been frequently used as a baseline for the other proposed methods.

The middle part of Table \ref{tab:recall} shows the results of the methods for mapping the local representations of images and languages, as shown in Subsection \ref{subsec:local}.
The overall retrieval accuracy is higher than that of the methods that map global representations of images and language.
This is because the words in the search query text can now be appropriately mapped to the objects in the image.
Among them, SCAN, which incorporates an attention mechanism, is highly accurate and has become the basis for various subsequent methods, such as BFAN, PFAN, and GSMN.

The bottom part of Table \ref{tab:recall} shows the results of the methods that utilize models trained on a large corpus of images and languages shown in Subsection \ref{subsec:pretrained}.
Note that in these methods, caption datasets other than Flickr30k and MSCOCO were used for pre-training.
The method of fine-tuning using the transformer-based pre-training model, which is powerful in the field of language processing, greatly exceeds the accuracy of conventional methods.

Note that the accuracies shown in Table \ref{tab:recall} are for methods that are trained or fine-tuned using the datasets of interest (Flickr30k and MSCOCO).
For this reason, CLIP, which is realized by complete zero-shot learning without using the targeted datasets, is excluded from Table \ref{tab:recall} because it is simply not comparable.
However, they show that its accuracy is almost as good as that of other fine-tuned transformer-based methods.

\section{Implementation}

The publicly available implementations that can be used to build the base system and compare the proposed method with the baseline method are summarized in Table \ref{tab:implementation}.
Among them, there are many improved source codes based on VSE++ and SCAN implemented using PyTorch.
The SCAN implementation is a modified version of the VSE++ implementation.
In SCAN, image features obtained from bottom-up attention implementation are used.
Subsequent technologies, such as PFAN, GSMN, and CVSE, are based on VSE++ or SCAN implementations and image features with bottom-up attention.
The deep learning frameworks used in both of these implementations are PyTorch.

However, few implementations of methods that use transformer-based pre-training models have been published.
UNITER provides the latest pre-trained models for download, which can be fine-tuned to tasks such as VQA, VCR, and image/text retrieval.
Similarly, for Oscar, pre-trained models can be downloaded and fine-tuned for the tasks of VQA, GQA, natural language visual reasoning for real (NLVR2), image/text retrieval, and image caption generation.
However, as of January 2021, the source code for the pre-training part has not been released.
For CLIP, it is relatively easy to implement zero-shot images or text retrieval because pre-trained models and sample source code for extracting features from images and text are available.

\section{Summary}

In this paper, we introduced recently proposed methods for zero-shot image retrieval, where the query is a sentence.
It is noteworthy that significant progress has been made in just over five years.
In particular, the use of transformer-based learning models, which have been rapidly introduced in the past year or two, is expected to expand in the future.

A future challenge is to reduce the high computational cost.
In particular, the training of transformer-based models is very time-consuming and requires a large amount of computing resources.
Training powerful models, such as the recently proposed GPT-3 \cite{GPT-3}, would cost at least \$4.6 million, and if a single GPU is used, it would theoretically take 355 years.
Therefore, it is increasingly important to improve the computation efficiency, including hardware.

\section*{Acknowledgements}

This work was partially supported by JSPS KAKENHI (Grant Number 18K11362). 


\end{document}